\pdfoutput=1

\documentclass[11pt]{article}

\usepackage{emnlp2021}

\usepackage{times}
\usepackage{latexsym}
\usepackage{xspace}
\usepackage{amsfonts}
\usepackage{amsmath}
\usepackage{inconsolata}

\usepackage{tabularx}
\usepackage{float}
\usepackage{algorithm}
\usepackage[noend]{algpseudocode}

\usepackage[T1]{fontenc}

\usepackage[utf8]{inputenc}

\usepackage{microtype}
\usepackage{dsfont}
\usepackage{pgfplots}
\pgfplotsset{compat=newest}
\usepgfplotslibrary{groupplots}
\usepgfplotslibrary{dateplot}
\makeatletter
\newcommand*{\centerfloat}{%
  \parindent \z@
  \leftskip \z@ \@plus 1fil \@minus \marginparwidth
  \rightskip \leftskip
  \parfillskip \z@skip}
\makeatother
\usepackage{xcolor}

\usepackage{booktabs}

\usepackage{todonotes}
\makeatletter
\newcommand*\iftodonotes{\if@todonotes@disabled\expandafter\@secondoftwo\else\expandafter\@firstoftwo\fi} 
\makeatother

\newcommand{\defn}[1]{\textbf{#1}}
\newcommand{\xx}{\mathbf{x}}
\newcommand{\yy}{\mathbf{y}}
\newcommand{\calY}{\mathcal{Y}}

\newcommand{\vocab}{\mathcal{V}}
\newcommand{\nmax}{{n_{\mathrm{max}}}}

\newcommand{\score}{\mathrm{score}}

\newcommand{\semdist}{\cos}

\newcommand{\eos}{\textsc{eos}\xspace}
\newcommand{\bos}{\textsc{bos}\xspace}
\newcommand{\abs}[1]{\lvert #1 \rvert}

\newcommand{\rpm}{\raisebox{.2ex}{$\scriptstyle\pm$}\,}

\usepackage{enumitem}

\title{A Plug-and-Play Method for Controlled Text Generation}

\newcommand{\ucambridge}{\normalfont \text{2}}
\newcommand{\ethz}{\text{\normalfont 1}}

\author{Dami\'an Pascual$^{\ethz}$ B\'eni Egressy$^{\ethz}$ Clara Meister$^{\ethz}$, Ryan Cotterell$^{\ethz,\ucambridge}$ Roger Wattenhofer$^{\ethz}$ \\
 $^{\ethz}$ETH Z\"{u}rich ~\;~ $^{\ucambridge}$University of Cambridge\\
  \href{mailto:damianp@ethz.ch}{\texttt{damianp@ethz.ch}}~\;~ \href{mailto:begressy@ethz.ch}{\texttt{begressy@ethz.ch}}~\;~ \href{mailto:clara.meister@inf.ethz.ch}{\texttt{clara.meister@inf.ethz.ch}} \\
  \href{mailto:ryan.cotterell@inf.ethz.ch}{\texttt{ryan.cotterell@inf.ethz.ch}}~\;~ \href{mailto:wattenhofer@ethz.ch}{\texttt{wattenhofer@ethz.ch}}
}

\begin{document}
\maketitle
\begin{abstract}
Large pre-trained language models have repeatedly shown their ability to produce fluent text. Yet even when starting from a prompt, generation can continue in many plausible directions. 
Current decoding methods with the goal of controlling generation, e.g., to ensure specific words are included, either require additional models or fine-tuning, or work poorly when the task at hand is semantically unconstrained, e.g., story generation.
In this work, we present a \textit{plug-and-play} decoding method for controlled language generation that is so simple and intuitive, it can be described in a single sentence: given a topic or keyword, we add a shift to the probability distribution over our vocabulary towards semantically similar words. We show how annealing this distribution can be used to impose \emph{hard} constraints on language generation, something no other plug-and-play method is currently able to do with SOTA language generators.
Despite the simplicity of this approach, we see it works incredibly well in practice: decoding from GPT-2 leads to diverse and fluent sentences while guaranteeing the appearance of given guide words. We perform two user studies, revealing that (1) our method outperforms competing methods in human evaluations; and (2) 
forcing the guide words to appear in the generated text has no impact on the fluency of the generated text.\footnote{Code: \url{https://github.com/dapascual/K2T}}\looseness=-1
\end{abstract}


\section{Introduction}
Having systems capable of automatically generating human-like text has been an objective pursued since the early days of artificial intelligence~\cite{meehan1977tale, lebowitz1987planning}. The recent development of large pre-trained language models based on the transformer architecture has brought us closer to this goal.
Indeed, current state-of-the-art models can produce impressively realistic text~\cite{gpt2,t5,gpt3}, facilitating their use across a wide range of language generation applications.\looseness=-1

Yet, these models are probabilistic in nature, and in certain language generation tasks, i.e., those that place few \defn{semantic constraints} on the output, it can be difficult to control the general theme or ensure the presence of specific words in generated text. We use story generation \cite{fan-etal-2018-hierarchical} as a running example; even in the presence of a prompt, there are many continuations that may result in a good story. Indeed, for this task, a generator that does not reflect this diversity is likely undesirable, since the number of stories it could produce would be limited. But what if you would like your story to fulfill certain criteria? Under existing methods, one would have to train (or at least fine-tune) a new model for a limited set of use-cases or hope that the model places sufficient probability mass on text meeting this criteria such that it shows up during search\footnote{Whilst this approach is often effective for semantically constrained tasks, such as machine translation, it becomes prohibitively expensive in practice for open-ended tasks.} \cite{hokamp2017,post2018,ziegler2019fine,keskar2019ctrl}.\looseness=-1

In this work, we introduce a better strategy for \defn{controlled decoding}---where generated text must contain certain words---for semantically unconstrained generation tasks. 
In short, we shift the output distribution of a language generation model towards the semantic space of a given guide word.
While this proposal sounds abstract at first, its realization is simple and intuitive: 
at each generation step, we modify the log-probability of the words in our vocabulary according to their semantic similarity to our guide word, quantified as the cosine similarity between their vector representations, i.e., word embeddings. 
This method not only encourages the explicit appearance of a guide word, but also encourages the model to generate appropriate context for the guide word to appear, i.e., words in the same semantic neighborhood \cite{hashimoto-etal-2016-word}. 
By annealing the strength of the probability shift, we can guarantee that all guide words appear in our text. Moreover, this method allows for guidance towards both ordered and unordered sets of words. 
Our decoding strategy is plug-and-play, i.e., it can be combined with any autoregressive language model and decoding algorithm with no additional training, differentiating it from other self-claimed plug-and-play methods that both (1) require additional discriminators and (2) lack the ability to ensure the appearance of specific words \cite{dathathri2019plug, krause2020gedi}. In particular, this means it is out-of-the-box compatible with large pre-trained transformer models~\cite{gpt3, t5}.

We run extensive experiments with GPT-2 as our language model and find that our method produces text containing all specified guide words without an impact on the fluency and diversity of generated text.
That is, we can control generation without harming text quality. Furthermore, we find that our strategy outperforms competing methods in human evaluations, and that generated text is close in fluency and overall quality to human text.

\section{Preliminaries}\label{sec:prelim}

\subsection{Semantic Spaces}
We model a semantic space as a vector space $\mathbb{R}^d$ over concepts where distances indicate semantic similarities \cite{hashimoto-etal-2016-word}.
In practice, this amounts to assigning a vector representation to each concept that reflects its semantic properties; arguably, the most basic unit that we can use to represent concepts are words \cite{pado-lapata-2003-constructing}. 
While the notion of a semantic space is abstract in nature, prior research suggests that word embedding algorithms may in fact provide rough representations of these spaces \cite{hashimoto-etal-2016-word}.\looseness=-1

Word embeddings, i.e., continuous word representations produced by algorithms like word2vec~\cite{word2vec} or GloVe~\cite{pennington-etal-2014-glove}, 
have been studied in depth. Notably, it has been shown that
cosine similarity between word embeddings learned by algorithms like GloVe provides a metric of semantic similarity \cite{erk-2009-representing,pennington-etal-2014-glove}. Motivated by these results, we adopt word embedding vectors $\gamma(w)$ as points representing concepts in our semantic space---where $\gamma: \vocab \rightarrow \mathbb{R}^d$ is the mapping learned by an embedding algorithm from a word in our vocabulary $w \in \vocab$ to the vector space $\mathbb{R}^d$; we adopt cosine similarity $\cos(\gamma(w), \gamma(w'))$ for words $w,w'$ as our notion of semantic similarity.

\subsection{Text Generation}

We consider probabilistic models $p$ that assign a probability to all sequences $\yy$ in the space of strings $\calY(\vocab, \nmax)$, where $\vocab$ is the model's vocabulary and $\nmax$ is the maximum sequence length considered;\footnote{Typically, we have $\calY(\vocab, \nmax) = \vocab^{\nmax}$, i.e., the set $\calY$ is exponentially large in $\vocab$.}
all sequences are padded with distinguished beginning-of- and end-of-sequence symbols \bos and \eos. 
In this work, we focus on the case of autoregressive locally normalized probabilistic models, for which the probability of a sequence $\yy$ can be decomposed using the chain rule of probability: \begin{equation}\label{joint-prob}
    p(\yy) = \prod_{t=1}^{|\yy|}p(y_t \mid \yy_{< t})
\end{equation}
\noindent In today's language generation tasks, $p$ is typically parameterized by a neural network, e.g., a transformer or a recurrent neural network; these models have led to impressive results, producing language that is both fluent and coherent~\cite{gpt2,t5,gpt3}.

Note that $p$ may additionally be conditioned on some input $\mathbf{x}$, e.g., an image or a sentence in a source language.\footnote{We leave the dependence on $\xx$ implicit when obvious from context.} Informally, we can view this conditioning as a shift of the mass of $p$ towards the semantic space a model learns to map to from input $\xx$; this shift may be strong, as in the case of machine translation, where the outcome should satisfy strict semantic constraints, or weak, as in topic-oriented story generation, where text should simply follow a general theme.\looseness=-1

The task of \defn{text generation} is to decode sequences of natural language, i.e., text, from $p$.
There are myriad strategies for decoding, with no single de facto method used for all language generation tasks. Yet with few exceptions, all can be described using a framework consisting of two components: a  \defn{score function} and a \defn{decoding algorithm}. Following the structure of Eq.~\ref{joint-prob}, we define decoding algorithms as the general class of algorithms---which may be stochastic or deterministic in nature---that decode text autoregressively according to a score function. Common examples include nucleus sampling \cite{nucleus} and beam search \cite{reddy-1977,meister+al.emnlp20}. 
Formally, we define a score function, $\score(\cdot\mid \yy_{<t})\!\!: \vocab^t \rightarrow \mathbb{R}$, as a map from strings generated under a model's vocabulary to a real number. As is clear from notation, we assume a dependency of this score on previously generated text $\yy_{<t} = \langle y_1, \dots, y_{t-1} \rangle$. For probabilistic text generators, the default score function is $\score(\cdot\mid \yy_{<t}) = \log p(\cdot\!\mid\!\yy_{<t})$.\footnote{Log-probability is often used for numerical stability.} Other examples include mutual information (only applicable in the presence of an input $\xx$): $\score(\cdot\mid \yy_{<t}) = \log (p(\cdot\!\mid\!\yy_{<t}, \xx)/p(\yy_{<t}))$  \cite{li-etal-2016-diversity}. We note that for sampling-based decoding algorithms, the distribution over $\vocab$ output by $\score(\cdot\mid \yy_{<t})$ is projected onto the $\Delta^{|\vocab|-1}$ probability simplex---typically via the softmax transformation---such that it can be sampled from.\footnote{
Algorithms that decode sets, like beam search, may additionally include a \emph{re-ranking} process with its own separate $\score$ function. This process allows a decoding algorithm to choose among a set of candidates, potentially taking into account global qualities of a sequence, such as length \cite{murray-chiang-2018-correcting} or coverage \cite{wu2016google}.}\looseness=-1

\subsection{Controlling Generation}

Two different types of control can be applied over language generation models: \defn{soft control} and \defn{hard control}. Soft control aims at directing the mood or the general topic of the generated text, whilst hard control aims at ensuring that some explicit constraints are met, e.g., specific words are contained in the text. Note that soft control can also be reached via hard control, i.e., text that contains a set of words related to a certain topic should arguably revolve around that topic.\looseness=-1

Some recent work has approached the problem of soft control on unconstrained language generation by training or fine-tuning language models \cite{ziegler2019fine,yu2017seqgan,keskar2019ctrl}. However, given the existing trend towards using out-of-the-box pre-trained language models, it is desirable to develop control methods that are plug-and-play, i.e., that can be applied on an existing model without additional training. Yet currently, even methods that are termed plug-and-play require the training of an external discriminator \cite{dathathri2019plug, krause2020gedi}.

While hard control of constrained generation, e.g., machine translation, can be attained with grid beam search methods~\cite{hokamp2017, post2018, hu2019improved}, it is impractical to use the same approach for hard control of unconstrained generation. Methods such as grid beam search rely on the assumption that there exists a core set of plausible candidates fulfilling the desired criteria.  While typically true for machine translation---where a well-calibrated model places most of its probability mass on a (relatively) small subset of natural language sequences---this is not often the case for open-ended generation tasks.
Recent work on stochastic search~\cite{miao2019cgmh, sha2020gradient} has approached this problem by performing bidirectional search during generation and editing the text until the constraints are fulfilled. Although stochastic search is suitable for bidirectional RNN models, it is not yet clear if it can be applied to forward generation models, e.g., transformer-based models.

\section{Keyword2Text} 

In this work, we propose \emph{Keyword2Text} (K2T), a new and simple plug-and-play method for exerting hard control during text generation. By modifying the score function, we can incorporate a semantic shift at decoding time, without additional models or fine-tuning. This method is model agnostic---it works with any autoregressive language model, including pre-trained transformers, and can be combined with other decoding methods and objectives. Further, we show that K2T (1) does not require a pre-defined ordering of constraints, and (2) can be used for guaranteeing hard constraints.\looseness=-1

\subsection{A Controlled Generation Objective}\label{sec:obj}

We consider a probabilistic language generator $p$     
and a word $w$, which is either a specific word we would like to appear in the generated text or a topic we would like to to steer generation towards. We refer to $\gamma(w) \in \mathbb{R}^d$ as our topic vector, i.e., the point in our semantic space associated with the word $w$. 
We propose a simple modification to the score function $\score(\cdot \mid \yy_{<t})$ to guide generation towards $w$:
\begin{align}\label{eq:score}
    \score'(y_{t}, &\,w \mid \yy_{<t}) = \score(y_{t}\mid \yy_{<t}) \\
    &\quad\quad + \lambda \cdot \max\Big( 0, \semdist(\gamma(y_{t}), \gamma(w))\Big) \nonumber
\end{align}
\noindent where $\lambda$ is a hyperparameter indicating the strength of the shift. As we take  $\lambda \rightarrow 0$, we recover our original score function; as $\lambda \rightarrow \infty$, the word $w$ will be assigned increasingly more weight, until it becomes the dominating term in Eq.~\ref{eq:score}.  

Note that we only consider positive similarities to avoid explicitly decrementing the scores of words that are favorable according to $\score(\cdot \mid \yy_{<t})$. Including negative cosine similarities could be used to control generation \emph{away} from certain words or concepts, e.g., text detoxification~\cite{dathathri2019plug,krause2020gedi}. 

\subsection{An Algorithm for Guided Generation}\label{sec:alg}
Using the framework given in \S\ref{sec:prelim}, the proposed score function can be used with any standard decoding algorithm, e.g., nucleus sampling or beam search. 
Yet incorporating this objective on its own will not guarantee that all the desired criteria for controlled decoding are met: it must be used in a principled way if we wish to work with multiple guide words or enforce the appearance of a word while still generating fluent text. We now present a general set of decoding algorithm modifications for controlled generation towards a list of guide words $W =[w_1, \dots, w_N]$.

\subsubsection{Ordered and Unordered Control}

Given a list of guide words $W$, we propose two approaches for both the case where we need words $w_n$ to appear in a fixed order as well as when any order suffices.\looseness=-1

\paragraph{Fixed Order.} 
We guide towards each of the words $w_n$ in turn, i.e. we start at $n=1$ and take $\score'(y_{t}, w_1\mid \yy_{<t})$ as our score function until the word $w_1$ appears in the generated text. 
Then, we switch to $\score'(y_{t}, w_2\mid \yy_{<t})$ until the word $w_2$ appears in the generated text.
We repeat this process until all $N$ words appear in the generated text.

\paragraph{Guide Closest.} We now treat $W$ as a set, i.e., the ordering of guide words is no longer important. At any given decoding step, we shift the score function by the highest cosine similarity across all words $w\in W_{t}$.\footnote{Note that we do not shift the tokens towards all guide words additively; this corresponds to shifting towards the mean of the guide word embeddings, which may correspond to a different area of the semantic space.}
Explicitly, we score $y_{t}$ as
\begin{align}
    \score'(y_{t}, W_{t}\mid \yy_{<t}) = & \,\score(y_{t} \mid \yy_{<t}) \ + \\
    \lambda \cdot \max &\Big(0, \underset{{w\in W_t}}{\max}\, \semdist(\gamma(y_t), \gamma(w)\Big) \nonumber
\end{align}

\noindent where we overload $\score'(\cdot\mid \yy_{<t})$ to take a set $W_t \subseteq W$---the guide words that have \emph{not} appeared before step $t$---as input.
Notably, this implies that the guide words do \emph{not} need to be ordered in our approach.
Previously proposed decoding algorithms, e.g. \citet{hokamp2017} and \citet{post2018}, run in exponential time without an ordering of the guide words.
Note that in either of our approaches discussed above, once all $w\in W$ appear in $\yy_{\leq t}$, we may revert back to the original score function $\score(\cdot \mid \yy_{<t})$.

\subsection{Guaranteeing Word Appearance}
We can guarantee the appearance of guide words when generating text by controlling the shift parameter $\lambda$ of Eq.~\ref{eq:score}.
As said, this parameter regulates the spectrum ranging from uncontrolled generation to forcing the next word to be a guide word. We propose to increase $\lambda$ on an exponential schedule. In words, as the generated sequence increases in length, so does the strength of the semantic shift, until we deterministically choose the guide word. \footnote{Note that if all the keywords have not appeared and the \emph{end-of-sequence} token is generated the algorithm discards it and samples again; i.e., \emph{end-of-sequence} is not allowed until all the keywords have been generated.}

Formally, suppose the maximum length of the sequence is $T$ and the previous guide word appeared at $t_n$ (we define $t_0=0$), then the weight at time $t$, where $t_n < t < T-\abs{W_t}$, is given by
\begin{equation}
\lambda_t =
\begin{cases}
     \lambda_0 \exp \left\{ \frac{c(t-t_n)}{T-\abs{W_t}-t_n} \right\}  & \textbf{if}\,\,t < T - \abs{W_t} \\
     \infty & \textbf{otherwise}
\end{cases}
\end{equation}

\noindent Thus, as we approach the maximum length, $T$, we exponentially increase the shift parameter. When we have only enough space for the remaining guide words, i.e. $t = T - \abs{W_t}$, we explicitly force the remaining guide words to appear by setting $\lambda_t = \infty$. We use the hyperparameters $\lambda_0$ and $c$ to control the initial value and growth of $\lambda$, respectively.\footnote{Empirically, we find that any reasonably large value of $c$ works well, e.g., $c=100$.}\looseness=-1

\section{Experimental Setup}\label{sec:exps}

To evaluate our controlled decoding strategy, we run three sets of experiments: (1) analysis of hyperparameters; (2) comparison to competing methods; (3) comparison to human text. In each of these experiments, the task consists of generating text that contains certain keywords. We use GPT-2 large \cite[$774$ million parameters;][]{gpt2} as our language model and GloVe as the embedding algorithm that generates $\gamma(\cdot)$. 
For experiments (2) and (3), we additionally run user studies; in all experiments we calculate the following automatic metrics, which in combination, we take as an (imperfect) estimate of text quality \cite{welleck2019neural,martins-etal-2020-sparse}.

\begin{itemize}[leftmargin=*]
\setlength\itemsep{0em}
    \item \emph{Perplexity}: exponentiation of the negative average per-token log-probability under a language model, lower is typically considered better. We use a separate model Distil-GPT-2~\cite{wolf2019huggingface} to calculate perplexity to avoid inflated estimates~\cite{liu2016not}.
    \item \emph{Repetition}: we use the sequence repetition score from \citet{welleck2019neural}, which computes the proportion of repeated 4-grams in the text; a lower score implies less repetitive text. As reference, \citet{welleck2019neural} find the repetition of human text from a subset of WikiText-103~\cite{merity2016pointer} to be $0.6\%$.
    \item \emph{Success Rate}: Percentage of total number of keywords that appear in the text. 
\end{itemize}

We use word stemming to check for the occurrence of the target words, i.e., if a generated word has the same stem as the keyword, it is counted as an occurrence; this way special cases such as plurals can be handled automatically, whilst keeping the text coherent by avoiding semantic redundancies, such as ``protective protection''.

\subsection{Hyperparameter Analysis}\label{sec:hyp}

To analyse hyperparameter choices, we devise a task that consists of generating a short piece of text that contains five keywords (cf. App.~\ref{app:exp_settings}).
We use solely the \emph{start of sequence} token as initial context for GPT-2 to start generating text, and we stop after $90$ tokens have been generated.
Since the keywords in each set bear no relation to each other, it is very challenging to include all of them in a coherent and fluent piece of text; in fact, generated samples tend to have high perplexity scores. Therefore, this task is highly demanding, making it a tough benchmark for controlled language generation.\looseness=-1 
\subsection{Comparison to Alternative Methods}\label{sec:compare}

Using the ROC story dataset, we compare against two methods for hard control in language generation, CGMH~\cite{miao2019cgmh} and Plan-and-write~\cite{yao2019plan}; as a trainable baseline we fine-tune GPT-2. See App.~\ref{app:exp_settings} for model details.\looseness=-1 

\paragraph{CGMH.} CGMH is a stochastic search method
that at each step samples a word in the generated sentence and an operation (replace, delete, insert). 
This method is plug-an-play but it works only with bidirectional language models. 
\citeauthor{miao2019cgmh} show that CGMH clearly outperforms grid beam search~\cite{hokamp2017} and the backward-forward model~\cite{mou2015backward}.\looseness=-1

\paragraph{Plan-and-write.} 
Plan-and-write is an end-to-end model for story generation, which takes a title as input and outputs a story. It proceeds in two steps: first, it generates a storyline represented by a sequence of keywords; then it generates a story based on the title and the storyline. 
\looseness=-1 

\paragraph{Fine-tuned GPT-2.} To fine-tune GPT-2, for each training example we extract five keywords $w_n$ from the story text using the publicly available YAKE library. Then, we provide these five keywords as a prompt together with the story text: ``\bos $w_1$, $w_2$, $w_3$, $w_4$, $w_5$ = <\emph{text}> \eos'', where \eos is the \emph{end-of-sequence} token. We fine-tune GPT2 on a Titan RTX GPU (24GB) for five epochs on causal language modeling, which takes around 26 hours. At evaluation time we give as initial context a prompt with the same format as during training and we use nucleus sampling with $p=0.9$.

\paragraph{}The objective of the ROC story dataset is to generate 
stories given a title.\footnote{We do not to evaluate on tasks like CommonGen~\cite{lin2020commongen} since texts for those tasks are expected to be very simple; thus performance metrics would not reveal how K2T affects aspects such as fluency, which are vital in the tasks we consider, e.g., generating stories from models like GPT-2.}
To conform the dataset to our task, i.e., keyword to text, we build a test set of $20$ titles randomly selected from the dataset and use the plan component of Plan-and-write \cite{yao2019plan} to generate five keywords from these titles.
Using these keywords we generate one story with each of the competing methods, resulting in $20$ examples per method. 
We perform a human evaluation comparing \emph{GPT-2+K2T}, \emph{CGMH} and \emph{Plan-and-write} in which $30$ evaluators are presented each of the three texts and asked to evaluate them relative to each other in terms of 4 criteria: fluency, logical consistency, creativity and best overall (see App.~\ref{app:exp_settings} and \ref{app:roc} for details).\looseness=-1

\subsection{Comparison to Human Text}

To assess the quality of text generated using K2T with respect to human text, we generate news articles from keywords. We employ the 500N-KPCrowd dataset from \citet{marujo2011keyphrase}, which consists of pieces of news written by professional journalists with keywords assigned by human annotators. For our evaluation, we randomly select ten keyword-article pairs from the test set.
At generation time, the language model receives the first $30$ words of the original article as initial context. In our baseline, the model receives no guidance, i.e., $\lambda=0$. We compare this to text generated from the same model albeit controlled by K2T.

We design our human evaluation based on best practices for evaluation of generated text \cite{van2019best}. 
Specifically, we prepare three sets of ten articles, each consisting of a combination of original articles (written by humans), articles generated by our method and articles generated by GPT-2 without control. We create three separate surveys so that evaluators will only be exposed to one version of each story (they do not know the origin). Participants are asked to evaluate how coherent, fluent and natural (human-like) each article is, as well as its overall quality, on a 7-point Likert scale~\cite{van2019best}, i.e., from $1$ to $7$ where higher is better (cf. App.~\ref{app:news}). Each survey is shown to $30$ evaluators.
In App.~\ref{app:news_texts} we show all the articles used in this study.\looseness=-1

\section{Results}

\subsection{Hyperparameter Analysis}

We perform the keyword to phrase task specified in \S\ref{sec:hyp}. We run each experiment ten times with different seeds and report the mean and standard deviation across the runs.
\begin{table}
\centering
\resizebox{\linewidth}{!}{%
\begin{tabular}{@{}lccc@{}} \toprule
 Method & SR (\%) & PPL & Rep. (\%)\\ 
 \midrule
 \emph{No control} & 0.6 \rpm 0.5 & 34.6 \rpm 3.2 & \textbf{2.4} \rpm 0.7\\ 
 \addlinespace[.5em]
 \emph{W.} $\lambda=5$ & 4.4 \rpm 0.9 & 34.5 \rpm 2.8 & 3.7 \rpm 0.8\\ 
 \emph{W.} $\lambda=10$ & 52.0 \rpm 3.4 & 46.7 \rpm 3.3 & 7.2 \rpm 1.3\\ 
 \emph{W.} $\lambda=20$ & 84.35 \rpm 1.2 & 225.9 \rpm 132.6 & 33.0 \rpm 1.5\\
 \addlinespace[.5em]
 \emph{C.} $\lambda=5$ & 12.2 \rpm 2.1 & \textbf{29.8} \rpm 1.3 & 3.3 \rpm 1.4\\
 \emph{C.} $\lambda=10$ & 72.6 \rpm 2.8 & 44.75 \rpm 3.7 & 8.7 \rpm 1.3\\
 \emph{C.} $\lambda=20$ & \textbf{95.1} \rpm 2.3 & 99.3 \rpm 20.1 & 13.4 \rpm 2.1\\ 
 \bottomrule
\end{tabular}
}
\caption{Comparison of the \emph{No control}, \emph{Guide words only (W)} and \emph{Guide Context (C)} strategies.}
\label{tab:logMod}
\end{table}
\paragraph{Controlled Generation.}

First, we assess the baseline effectiveness of our generation objective in Eq.~\ref{eq:score} for generating text that contains a set of guide words $W$. We compare three approaches.\looseness=-1

\begin{itemize}[leftmargin=*]
\setlength\itemsep{0em}
    \item \emph{No control}: language generation without guidance, i.e., $\lambda=0$;
    \item \emph{Guide words only}:\footnote{This is similar to the weighted decoding (WD) baseline used in \cite{dathathri2019plug}.} shifting the scores only for the tokens corresponding exactly to the guide words $W$ and not for similar words;
    \item \emph{Guide context}: shifting the score for the guide words $W$ in addition to semantically similar words. 
\end{itemize}

In this evaluation we do not anneal $\lambda$; thus, guide words are \emph{not} guaranteed to appear.
It follows that we can only evaluate the performance of our method under soft constraints. 
Unless otherwise stated, here and in the following, we apply unordered control with the \emph{Guide Closest} strategy, we use $\score(\cdot\mid \yy_{<t}) = \log p(\cdot\!\mid\!\yy_{<t})$ as our original score function and we use nucleus sampling with $p=0.9$ as the decoding algorithm.
We present the results of this comparison in Table~\ref{tab:logMod}. 

When the score function is not shifted (\emph{No control}), the average success rate is only $0.6\%$, which serves as a random baseline. We see that guiding towards the context is more effective than guiding only towards the exact guide words, i.e., for the three values of $\lambda$ considered here, both success rate and perplexity are significantly better.  
Furthermore, the repetition score of \emph{Guide Context} is clearly better for $\lambda=20$, the only value where both approaches reach a high success rate. Note that encouraging the appearance of specific keywords also helps keep the generated text on-topic; this may explain the lower perplexity of the best performing configuration versus \emph{No control}.
These results validate \emph{Guide Context} as the best strategy. 

\paragraph{Initial Shift Strength $\lambda_0$.}
Next, we investigate the effect of varying the initial shift strength $\lambda_0$; here and in the remaining experiments we guarantee the appearance of all keywords. 
In Table~\ref{tab:logStr} we report perplexity and repetition for different values of $\lambda_0$ (success rate is 100\% in all cases).
\begin{table}
\centering
\begin{tabular}{@{}lcc@{}} \toprule
 $\lambda_0$ & PPL & Rep. (\%) \\
 \midrule
 5 & \textbf{58.4} \rpm 4.5 & \textbf{3.5} \rpm 1.1 \\
 10 & 70.5 \rpm 7.1 & 6.4 \rpm 2.2 \\ 
 15 & 109.5 \rpm 24.2 & 10.5 \rpm 3.2 \\
 20 & 235.8 \rpm 352.2 & 10.6 \rpm 1.7 \\
 25 & 135.8 \rpm 44.9 & 9.9 \rpm 2.4 \\
 30 & 310.3 \rpm 366.8 & 9.5 \rpm 2.1  \\
 \bottomrule
\end{tabular}
\caption{Evaluation of the shift strength $\lambda$.}
\label{tab:logStr}
\end{table}
We see that increasing $\lambda_0$ results in a worsening of perplexity and repetition scores, with strong variability across different runs, i.e., high standard deviation. For $\lambda_0=5$ the average perplexity ($58.4$) and repetition score ($3.5\%$) are the best among the considered values; we use this value of $\lambda_0$ in the remaining experiments.

\paragraph{Unordered Control.}

Given a set of words, it is not trivial to devise a strategy to guide text generation towards those words without a pre-specified order. We explore two additional strategies on top of \emph{Guide Closest}, described in \S\ref{sec:alg}:

\begin{itemize}[leftmargin=*]
\setlength\itemsep{0em}
    \item \emph{Guide All}: We shift the scores towards all guide words at once, by adding the sum of the cosine similarities. 
    \item \emph{Guide Random}: At each generation step we choose the next guide word uniformly at random from the remaining guide words and shift the scores towards this word.
\end{itemize}

In Table~\ref{tab:unordered} we present the evaluation of these strategies. For reference, we also report results for guiding the guide words in order. We see that guiding towards \emph{all} the words at the same time performs poorly in terms of repetition score ($30.0\%$). 
On the other hand, \emph{Guide Random} and \emph{Guide Closest} produce similar results, with smaller perplexity for the \emph{Guide Closest} strategy and thus, in the following we adopt the \emph{Guide Closest} strategy.

\begin{table}
\centering
\begin{tabular}{@{}lcc@{}} \toprule
 Strategy & PPL & Rep. (\%) \\
 \midrule
 \emph{Guide Closest} & 58.4 \rpm 4.5 & 3.5 \rpm 1.1\\
 \emph{Guide All} & \textbf{39.7} \rpm 2.7 & 30.0 \rpm 2.3\\
 \emph{Guide Random} & 66.9 \rpm 3.7 & \textbf{1.5} \rpm 0.5 \\
 \addlinespace[.5em]
 \emph{Fixed Order} & 61.7 \rpm 4.2 & 3.4 \rpm 1.2 \\
 \bottomrule
\end{tabular}
\caption{Comparison of different approaches for control towards unordered constraints with $\lambda_0=5$.}
\label{tab:unordered}
\end{table}

\paragraph{Decoding Algorithm.}

Finally, we evaluate our method in conjunction with different decoding algorithms:\looseness=-1 

\begin{itemize}[leftmargin=*]
\setlength\itemsep{0em}
    \item Nucleus sampling \cite[NS;][]{nucleus} with $p=0.9$. 
    \item Beam search (BS) with $4$ beams and length normalized re-ranking, i.e. taking the top-$K$ candidates according to $\score(\yy) = \log p(\yy)/|\yy|$.\looseness=-1
    \item BS with word count (BS+WC): we increment the above re-ranking function if the generated token $y_t$ corresponds to one of the remaining guide words:
    \begin{align}
    \quad \score'(&\yy_{< t}) = \score(\yy_{< t}) \\
    &\quad\quad\quad\quad + \Big(\abs{W} - \abs{W_{t+1}}\Big) \nonumber
    \end{align}
    where $\abs{W} - \abs{W_t} \geq 0$ equals the number of guide words that have appeared by step $t$.
    \item BS+WC with NS (BS+WC+NS): we consider a hybrid of beam search and nucleus sampling where the words are sampled using nucleus (instead of deterministically picking the top words). 
\end{itemize}

\begin{table}
\centering
\begin{tabular}{@{}lcc@{}} \toprule
 Strategy & PPL & Rep. (\%)\\
 \midrule
 \emph{NS} & 58.4 \rpm 4.5 & \textbf{3.5} \rpm 1.1 \\
 \emph{BS} & \textbf{9.8} \rpm 0 & 46.5 \rpm 0 \\
 \emph{BS+WC} & 11.9 \rpm 0 & 38.7 \rpm 0 \\
 \emph{BS+WC+NS} & 21.2 \rpm 1.7 & 13.4 \rpm 2.2 \\
 \bottomrule
\end{tabular}
\caption{Comparison of different decoding algorithms.}
\label{tab:dec}
\end{table}

Table~\ref{tab:dec} shows that while NS produces diverse text, its perplexity is the highest by a margin. Conversely, BS with and without word count reward, generates text with low perplexity but high repetition. 
The combination of both reaches the best trade-off.\looseness=-1

\renewcommand{\arraystretch}{1.3}
\begin{table}
\centering
\begin{tabular}{@{}p{\columnwidth}@{}} \toprule
Words: \textit{Guide, Jump, Eight, Row, Settle} \\
\addlinespace[.2em]
Text: \textit{One of the most common questions I get from new \textbf{rowers} is how to warm up before a race. \textbf{Eight} months ago, I wrote a \textbf{guide} to warm up and how to \textbf{jump} in the water, but I never got around to updating it. So I decided to \textbf{settle} the question once and for all, and try to write the best possible course through it.}
\\
\midrule
Words: \textit{Search, Major, String, Cost, Village} \\
\addlinespace[.2em]
Text: \textit{A \textbf{major} \textbf{string} of high-profile \textbf{cost}-cutting measures announced by the Ontario government this week is expected to result in a significant drop in the \textbf{village}'s operating budget by the end of the year. The \textbf{search} for ways to lower the village's pension plan contribution from 7.5 per cent to 6.9 per cent is expected to save an additional \$80 million in the current fiscal year.}\\
\bottomrule
\end{tabular}
\caption{Examples of text generated by K2T.}
\label{tab:ex}
\end{table}
\renewcommand{\arraystretch}{1}
In Table~\ref{tab:ex} we show two examples generated with the best-performing configuration: $\lambda_0=5$, \emph{Guide Closest}, \emph{BS+WC+NS} decoding, which we use for the experiments in \S\ref{sec:ROC} and \S\ref{sec:News}.

\subsection{Comparison to Alternative Methods}
\label{sec:ROC}

\begin{table}
\centering
\resizebox{\linewidth}{!}{%
\begin{tabular}{@{}lccc@{}} \toprule
 Method & SR (\%) & PPL & Rep. (\%)\\ 
 \midrule
 \emph{Plan-and-Write} & 96.0 & \textbf{33.9} & 25.7 \\ 
 \emph{CGMH} & 97.0 & 127.8 & 1.6 \\ 
 \emph{GPT-2 fine-tuned} & 72.0 & 89.4 & 1.8 \\
 \emph{GPT-2 + K2T} & \textbf{100.0} & 48.8 & \textbf{1.5} \\
 \bottomrule
\end{tabular}}
\caption{Comparison to competing methods.}
\label{tab:ROC_stats}
\end{table}

In Table \ref{tab:ROC_stats}, we report metrics for each of the three methods (K2T, Plan-and-write and CGMH) on the keyword-to-text task based on the ROC dataset. 
Neither Plan-and-write (P\&W), CGMH nor fine-tuned GPT-2 reach a $100\%$ success rate and, while P\&W presents the lowest perplexity, its repetition score is by far the highest ($25.7$). Conversely, although the repetition score of CGMH and GPT-2 fine-tuned are comparable to our method, they generate text with very high perplexity, $127.8$ and $89.4$ respectively. In light of these results, K2T seems to provide the best trade-off between perplexity and repetition, on top of reaching $100\%$ success rate.\looseness=-1

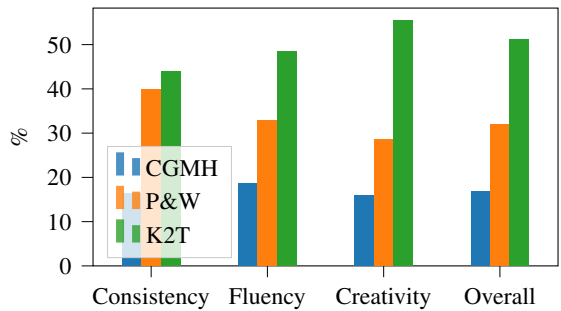
\begin{figure}
\begin{tikzpicture}

\definecolor{color0}{rgb}{0.12156862745098,0.466666666666667,0.705882352941177}
\definecolor{color1}{rgb}{1,0.498039215686275,0.0549019607843137}
\definecolor{color2}{rgb}{0.172549019607843,0.627450980392157,0.172549019607843}

\begin{axis}[
legend cell align={left},
legend style={at={(0.03,0.03)}, anchor=south west, fill opacity=0.8, draw opacity=1, text opacity=1, draw=white!80!black, font=\small},
tick align=outside,
tick pos=left,
x grid style={white!69.01960784313725!black},
xmin=-0.5, xmax=3.5,
xtick style={color=black},
xtick={0,1,2,3},
xticklabels={Consistency,Fluency,Creativity,Overall},
y grid style={white!69.01960784313725!black},
ylabel={\%},
ymin=0, ymax=0.58258064516129,
ytick style={color=black},
ytick={0,0.1,0.2,0.3,0.4,0.5},
yticklabels={0,10,20,30,40,50},
height = {5.0cm},
width = {1.0\linewidth},
tick label style={font=\small},
label style={font=\small}
]
\draw[fill=color0,draw opacity=0] (axis cs:-0.25,0) rectangle (axis cs:-0.0833333333333333,0.162903225806452);
\addlegendimage{ybar,ybar legend,fill=color0,draw opacity=0};
\addlegendentry{CGMH}

\draw[fill=color0,draw opacity=0] (axis cs:0.75,0) rectangle (axis cs:0.916666666666667,0.187096774193548);
\draw[fill=color0,draw opacity=0] (axis cs:1.75,0) rectangle (axis cs:1.91666666666667,0.159677419354839);
\draw[fill=color0,draw opacity=0] (axis cs:2.75,0) rectangle (axis cs:2.91666666666667,0.167741935483871);
\draw[fill=color1,draw opacity=0] (axis cs:-0.0833333333333333,0) rectangle (axis cs:0.0833333333333333,0.398387096774194);
\addlegendimage{ybar,ybar legend,fill=color1,draw opacity=0};
\addlegendentry{P\&W}

\draw[fill=color1,draw opacity=0] (axis cs:0.916666666666667,0) rectangle (axis cs:1.08333333333333,0.329032258064516);
\draw[fill=color1,draw opacity=0] (axis cs:1.91666666666667,0) rectangle (axis cs:2.08333333333333,0.285483870967742);
\draw[fill=color1,draw opacity=0] (axis cs:2.91666666666667,0) rectangle (axis cs:3.08333333333333,0.319354838709677);
\draw[fill=color2,draw opacity=0] (axis cs:0.0833333333333333,0) rectangle (axis cs:0.25,0.438709677419355);
\addlegendimage{ybar,ybar legend,fill=color2,draw opacity=0};
\addlegendentry{K2T}

\draw[fill=color2,draw opacity=0] (axis cs:1.08333333333333,0) rectangle (axis cs:1.25,0.483870967741935);
\draw[fill=color2,draw opacity=0] (axis cs:2.08333333333333,0) rectangle (axis cs:2.25,0.554838709677419);
\draw[fill=color2,draw opacity=0] (axis cs:3.08333333333333,0) rectangle (axis cs:3.25,0.512903225806452);
\end{axis}

\end{tikzpicture}
\caption{User study comparing the three methods; bars indicate how often each method was picked as the best.}
\label{fig:ROC_results}
\end{figure}

In Figure \ref{fig:ROC_results} we see the results of the user study comparing CGMH, P\&W and GPT-2+K2T. For each method, we report the proportion of times that it is chosen as the best of the three across four different axes: fluency, consistency, creativity and overall quality. K2T outperforms the other two in all aspects, in particular in creativity and overall quality. 
These results suggest that our method is superior to the baseline hard control methods on both objective and subjective metrics.

\subsection{Comparison to Human Text}
\label{sec:News}

In Table \ref{tab:News_stats} we report perplexity, repetition score and success rate on the news article task. We compare the original news articles with those generated by uncontrolled GPT-2 and with GPT-2 controlled by K2T. 
We see that without controlling GPT-2, the keywords do not appear in the generated texts, despite the $30$ words of initial context. This suggests that methods such as grid beam search \cite{hokamp2017,post2018} would perform poorly. We also see that when GPT-2 is controlled by K2T, the text it generates is close to the original text in terms of perplexity and repetition score.
To explore the perceived impact of the differences, we look at the results of the human evaluations.

\begin{table}
\centering
\begin{tabular}{@{}lccc@{}} \toprule
 Text & SR (\%) & PPL & Rep. (\%)\\ 
 \midrule
 \emph{Original \hspace{1cm}} & 100.0 & 15.2 & 1.3 \\ 
 \emph{GPT-2} & 0.0 & 8.8 & 11.5 \\ 
 \emph{Ours} & 100.0 & 12.5 & 1.0 \\ 
 \bottomrule
\end{tabular}
\caption{Comparison of the \emph{original} articles and text generated by \emph{GPT-2} without and with control (\emph{ours}).}
\label{tab:News_stats}
\end{table}

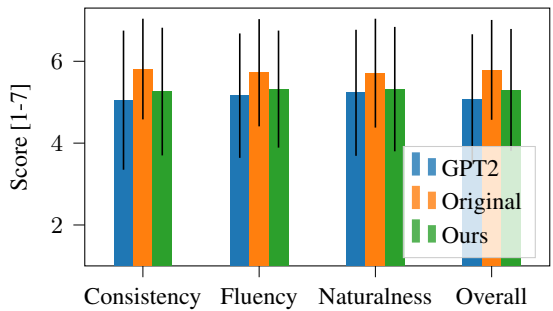
\begin{figure}
\begin{tikzpicture}

\definecolor{color0}{rgb}{0.12156862745098,0.466666666666667,0.705882352941177}
\definecolor{color1}{rgb}{1,0.498039215686275,0.0549019607843137}
\definecolor{color2}{rgb}{0.172549019607843,0.627450980392157,0.172549019607843}

\begin{axis}[
legend cell align={left},
legend style={at={(0.97,0.03)}, anchor=south east, fill opacity=0.8, draw opacity=1, text opacity=1, draw=white!80!black, font=\small},
tick align=outside,
tick pos=left,
x grid style={white!69.01960784313725!black},
xmin=-0.5, xmax=3.5,
xtick style={color=black},
xtick={0,1,2,3},
xticklabels={Consistency,Fluency,Naturalness,Overall},
y grid style={white!69.01960784313725!black},
ylabel={Score [1-7]},
ymin=1, ymax=7.3,
ytick style={color=black},
height = {5.0cm},
width = {1.0\linewidth},
tick label style={font=\small},
label style={font=\small}
]
\draw[fill=color0,draw opacity=0] (axis cs:-0.25,0) rectangle (axis cs:-0.0833333333333333,5.05);
\addlegendimage{ybar,ybar legend,fill=color0,draw opacity=0};
\addlegendentry{GPT2}

\draw[fill=color0,draw opacity=0] (axis cs:0.75,0) rectangle (axis cs:0.916666666666667,5.16);
\draw[fill=color0,draw opacity=0] (axis cs:1.75,0) rectangle (axis cs:1.91666666666667,5.23);
\draw[fill=color0,draw opacity=0] (axis cs:2.75,0) rectangle (axis cs:2.91666666666667,5.08);
\draw[fill=color1,draw opacity=0] (axis cs:-0.0833333333333333,0) rectangle (axis cs:0.0833333333333333,5.81);
\addlegendimage{ybar,ybar legend,fill=color1,draw opacity=0};
\addlegendentry{Original}

\draw[fill=color1,draw opacity=0] (axis cs:0.916666666666667,0) rectangle (axis cs:1.08333333333333,5.72);
\draw[fill=color1,draw opacity=0] (axis cs:1.91666666666667,0) rectangle (axis cs:2.08333333333333,5.71);
\draw[fill=color1,draw opacity=0] (axis cs:2.91666666666667,0) rectangle (axis cs:3.08333333333333,5.79);
\draw[fill=color2,draw opacity=0] (axis cs:0.0833333333333333,0) rectangle (axis cs:0.25,5.26);
\addlegendimage{ybar,ybar legend,fill=color2,draw opacity=0};
\addlegendentry{Ours}

\draw[fill=color2,draw opacity=0] (axis cs:1.08333333333333,0) rectangle (axis cs:1.25,5.32);
\draw[fill=color2,draw opacity=0] (axis cs:2.08333333333333,0) rectangle (axis cs:2.25,5.32);
\draw[fill=color2,draw opacity=0] (axis cs:3.08333333333333,0) rectangle (axis cs:3.25,5.3);
\path [draw=black, semithick]
(axis cs:-0.166666666666667,3.35)
--(axis cs:-0.166666666666667,6.75);

\path [draw=black, semithick]
(axis cs:0.833333333333333,3.64)
--(axis cs:0.833333333333333,6.68);

\path [draw=black, semithick]
(axis cs:1.83333333333333,3.69)
--(axis cs:1.83333333333333,6.77);

\path [draw=black, semithick]
(axis cs:2.83333333333333,3.5)
--(axis cs:2.83333333333333,6.66);

\path [draw=black, semithick]
(axis cs:0,4.58)
--(axis cs:0,7.04);

\path [draw=black, semithick]
(axis cs:1,4.41)
--(axis cs:1,7.03);

\path [draw=black, semithick]
(axis cs:2,4.38)
--(axis cs:2,7.04);

\path [draw=black, semithick]
(axis cs:3,4.57)
--(axis cs:3,7.01);

\path [draw=black, semithick]
(axis cs:0.166666666666667,3.7)
--(axis cs:0.166666666666667,6.82);

\path [draw=black, semithick]
(axis cs:1.16666666666667,3.89)
--(axis cs:1.16666666666667,6.75);

\path [draw=black, semithick]
(axis cs:2.16666666666667,3.8)
--(axis cs:2.16666666666667,6.84);

\path [draw=black, semithick]
(axis cs:3.16666666666667,3.81)
--(axis cs:3.16666666666667,6.79);

\end{axis}

\end{tikzpicture}
\caption{User study comparing our control method to \emph{original} and \emph{GPT-2} articles.}
\label{fig:News_results}
\end{figure}

Figure~\ref{fig:News_results} shows the results of the human evaluations on the news article task. As expected, news articles written by professional journalists are assessed superior to those written by GPT-2 when controlled by K2T. Although the difference is statistically significant ($p < 0.05$; $t$-test), it is small: less than $0.6$ points for each of the evaluated properties on a scale from $1$ to $7$. Furthermore, the text generated when controlling GPT-2 is comparable in terms of naturalness and fluency, and significantly better ($p < 0.05$; $t$-test) in terms of consistency and overall quality, than the text from uncontrolled GPT-2.
These results demonstrate that K2T does not compromise the quality of the generated text. Further, the resulting text is close in perceived quality to human text.

\section{Related Work}
Various other approaches exist for controlled language generation; here we review those that are not discussed in \S\ref{sec:prelim}.

\paragraph{Hard Control.} Hard control of autoregressive language models in unconstrained settings has remained elusive thus far. \citet{xu2020megatron} propose a framework to exert hard control over language generation which, nevertheless, requires training three large transformer models.

\paragraph{Soft Control.} 
Related to our approach of using semantic spaces to control generation, \citet{chang2021changing} use the space of GloVe embeddings to define control topics.
However, this method is not plug-and-play since it requires fine-tuning a GPT-2 encoder to generate text aligned with the control topics. A number of plug-and-play methods for soft control exist that use external discriminators to steer language generation. 
\citet{holtzman2018learning} train discriminators on different linguistic properties to improve the quality of generated text. 
\citet{dathathri2019plug} use the gradients of an external discriminator to direct the generation of a pre-trained language model towards the target topic. Similarly, ~\citet{krause2020gedi} use a contrastive strategy to soft-control language generation. \citet{yang2021fudge} directly modify the output probabilities of a (pre-trained) language model using the output of a discriminator that determines whether future text will contain the desired attribute, e.g., formality. 
Unfortunately, the use of external discriminators limits the applicability of these methods since they require training data for each of the target topics or attributes. Our method elegantly dispenses with the need for a discriminator by using the geometric properties of semantic spaces.

\vspace{-.5cm}
\section{Conclusion}

In this work, we present K2T, a simple and intuitive plug-and-play decoding method that can be used to impose controls on any autoregressive model for language generation, including large pre-trained transformers, like GPT-2. Our method guarantees the appearance of guide words and requires neither re-training nor the use of external discriminators. 
Our two user studies reveal that K2T is superior to competing hard control methods and that there is no statistical difference in perceived fluency, consistency and overall quality between news articles generated by our method and by GPT-2 without control. In future work, we plan to investigate in more detail the application of this decoding method to soft control tasks and to text detoxification.\looseness=-1

\section*{Ethics Statement}

We recognize that controllable language generation can potentially be used to produce misinformation or offensive text. However, we believe that further research on controllable generation is necessary to ensure that we have at our disposal the tools needed to prevent automatic language generation techniques from being used for malicious purposes. 
Methods for controlling large pre-trained language models are also a promising tool to mitigate the generation of biased text, e.g., steering generation towards the semantic space of both ``woman'' and ``doctor'' may mitigate the bias typically seen surrounding stereotypically male-oriented professions. We believe this is an important future direction for our work.
Further, we consider the environmental impacts of our method. As our method is completely plug-and-play, i.e., it is able to make use of pre-trained language models and word embeddings, we hope that it can be utilized in place of techniques that require extensive model training. This would in turn reduce the energy consumption required to setup a controllable language generation system, which for large language generation models, can be quite significant.

\bibliography{anthology,custom}
\bibliographystyle{acl_natbib}

\appendix
\onecolumn
\clearpage
\section{Evaluation Details}\label{app:exp_settings}

In this appendix we provide additional details on different aspects of our evaluation to ease reproducibility. 

\subsection{Keywords for Hyperparameter Analysis}

To generate the keyword sets we use in our hyperparameter analysis we use a list of $1000$ common English words\footnote{\url{https://github.com/powerlanguage/word-lists}}. From this list, we discard the first $500$ words, which we found to be too common and from the remaining $500$ words, we additionally discard stop words. Then, from the resulting list we sample $50$ sets of $5$ words each, which constitutes our keyword sets.

\subsection{Details of Alternative Methods}

For CGMH \cite{miao2019cgmh}, we use the same model as in the original work: a two-layer LSTM (63M parameters); and train it using a TITAN Xp 12GB GPU following the instructions from the original repository\footnote{\url{https://github.com/NingMiao/CGMH}}. For generating the samples, we run the model for $1000$ updates with a minimum sequence length of $30$ words. 
Similarly, Plan-and-write \cite{yao2019plan} uses an LSTM based sequence-to-sequence model with 62M parameters. We train the model for $357$ epochs on a TITAN Xp 12GB GPU with the ROC story dataset~\cite{mostafazadeh2016corpus}.

\subsection{Human Evaluation}
We perform both user studies using Amazon MTurks. In each of them, we include simple comprehension questions as a control to ensure that the evaluators read the texts carefully. We discard and replace those evaluators who do not spend the minimum expected time on the study or who do not answer enough control questions correctly.

\subsection{Comparison to Human Text}

To generate the news articles using the 500N-KPCrowd dataset from \citet{marujo2011keyphrase}, we randomly select ten keyword-article pairs from the test set with a length ranging from $81$ to $217$ words (so that the articles are neither too short nor too long for human assessment). Then, from the given keywords, we first remove proper nouns, e.g., names and cities. As the number of keywords provided in some articles is quite large, we further pare this list down by randomly removing selected keywords until the proportion is one keyword per ten words of text, such that all article-keyword pairs have present the same ratio of keywords per length.\looseness=-1

\clearpage
\section{ROC Story Survey}
\label{app:roc}
Here we provide a screenshot to show how our survey for method comparison looks.

\begin{figure}
    \centering
    \includegraphics[width=0.8\textwidth]{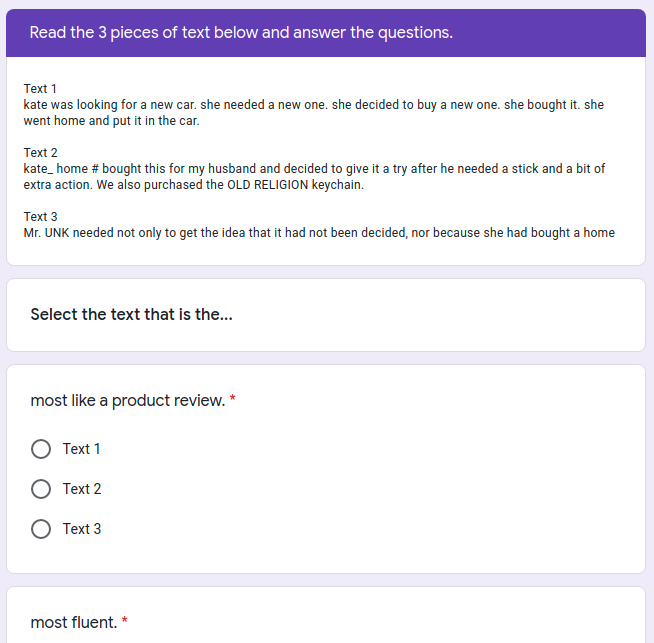}
    \caption{Screenshot of the ROC story user study.}
    \label{fig:roc_screenshot}
\end{figure}

\section{News Articles}\label{app:news}

\subsection{Survey}
As above, here we provide a screenshot that shows how our survey for comparison to human text looks.

\begin{figure}
    \centering
    \includegraphics[width=0.8\textwidth]{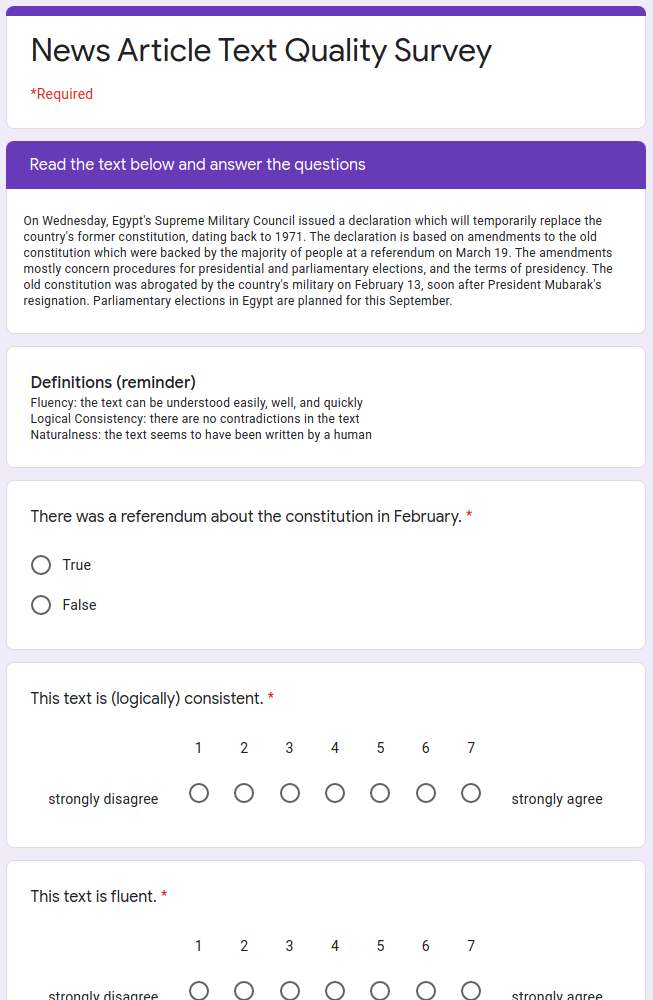}
    \caption{Screenshot of the news article user study.}
    \label{fig:news_screenshot}
\end{figure}

\newpage
\subsection{Texts}
\label{app:news_texts}

Below we show the original news articles and the news articles generated by our method that were used in the human evaluation study. Detailed results of the study are included in Appendix \ref{app:news_results}. 

\renewcommand{\arraystretch}{1.3}

\begin{center}
\begin{tabularx}{\linewidth}{X}
\hline
\hline
Text 1 \\ 
Context: \textcolor{blue}{Johnny Depp will be asked to make a fifth Pirates Of The Caribbean film if the fourth instalment is a success. Producer Jerry Bruckheimer said he already has a screenplay} \\ 
Keywords: fifth, audience, embraces, character, original, trilogy, continue, story, digital, cameras, pirates, fresh, new, carry \\
\hline
\textsc{Ours + Gpt2-774M}: \textit{Johnny Depp will be asked to make a fifth Pirates Of The Caribbean film if the fourth instalment is a success. Producer Jerry Bruckheimer said he already has a screenplay with \textbf{Pirates} \textbf{new} \textbf{trilogy} \textbf{character} Johnny Depp and \textbf{digital} animation studio Story Studio. The \textbf{fifth} film will \textbf{continue} the \textbf{original} \textbf{story} from the first three films and will be set in the Caribbean.
"We are very excited to \textbf{carry} on the story of the \textbf{audience}'s favourite Pirates of the Caribbean characters," said Bruckheimer. "We are \textbf{fresh} off the success of Pirates of the Caribbean: On Stranger Tides and we are looking forward to \textbf{embracing} the new generation of fans with a new film that will be even more exciting than the first three films."
The Pirates of the Caribbean: On Stranger Tides \textbf{cameras} have been filming in Barranquilla since April 11. Filming will continue for two months to reach completion by May 23.}\\
\hline
\textsc{Human}: \textit{Johnny Depp will be asked to make a fifth Pirates Of The Caribbean film if the fourth instalment is a success. Producer Jerry Bruckheimer said he already has a screenplay in the works for a \textbf{fifth} Pirates tale, which would follow the forthcoming Pirates Of The Caribbean: On Stranger Tides. ``As long as the \textbf{audience} \textbf{embraces} this one, we'll certainly try to make another one. It's really up to Johnny. He loves the \textbf{character},'' he said. The \textbf{original} three Pirates blockbusters ended up as a \textbf{trilogy} \textbf{continuing} the same key characters and \textbf{story} line. But Jerry said On Stranger Tides - the first Pirates flick shot with \textbf{digital} 3D \textbf{cameras} - and future \textbf{Pirates} films will be stand-alone stories continuing the adventures of Johnny's woozy buccaneer Captain Jack Sparrow. "The audience told us (at test screenings of On Stranger Tides) what they loved about it is that it was \textbf{fresh}, it was new, it was a whole \textbf{new} story, ``Jerry said. "So that will \textbf{carry} over into the next one, too, to give it something fresh and different.''} 
\\
\hline
\hline
\end{tabularx}
\end{center}
\clearpage

\begin{center}
\begin{tabularx}{\linewidth}{X}
\hline
\hline
Text 2 \\ 
Context: \textcolor{blue}{On Wednesday, Egypt's Supreme Military Council issued a declaration which will temporarily replace the country's former constitution, dating back to 1971. The declaration is based on amendments to the old} \\ 
Keywords: constitution, majority, elections, military, resignation \\
\hline
\textsc{Ours + Gpt2-774M}: \textit{On Wednesday, Egypt's Supreme Military Council issued a declaration which will temporarily replace the country's former constitution, dating back to 1971. The declaration is based on amendments to the old \textbf{constitution} which were approved by the \textbf{majority} of \textbf{military} members in \textbf{elections} earlier this year.
The new constitution will replace the previous one, which was approved by a referendum in June. The \textbf{resignation} of President Mohammed Morsi in July 2011 was a prerequisite for the current constitutional amendment.
Civil society activists point to the judiciary, a particularly divisive institution. They accuse the military council and authorities of targeting the judiciary to weaken the role of the Egyptian people.}\\
\hline
\textsc{Human}: \textit{On Wednesday, Egypt's Supreme Military Council issued a declaration which will temporarily replace the country's former constitution, dating back to 1971. The declaration is based on amendments to the old \textbf{constitution} which were backed by the \textbf{majority} of people at a referendum on March 19. The amendments mostly concern procedures for presidential and parliamentary \textbf{elections}, and the terms of presidency. The old constitution was abrogated by the country's \textbf{military} on February 13, soon after President Mubarak's \textbf{resignation}. Parliamentary elections in Egypt are planned for this September.} 
\\
\hline
\hline
\end{tabularx}
\end{center}

\begin{center}
\begin{tabularx}{\linewidth}{X}
\hline
\hline
Text 3 \\ 
Context: \textcolor{blue}{A group of moderate House Democrats have initiated conversations with Republican lawmakers in a bid to try to reach a deal to tackle burgeoning federal spending. The group is being} \\ 
Keywords: leaders, coalition, conservative, lawmakers, goals \\
\hline
\textsc{Ours + Gpt2-774M}: \textit{A group of moderate House Democrats have initiated conversations with Republican lawmakers in a bid to try to reach a deal to tackle burgeoning federal spending. The group is being led by \textbf{conservative} \textbf{lawmakers} and \textbf{leaders} of the \textbf{coalition}, including Reps. Chris Van Hollen (Md.), Jim McGovern (Mass.), and Jim McDermott (Wash.). The \textbf{goals} are to present a united front and try to provide some bipartisan credibility to the nascent spending talks.
That effort will likely fail, however, as the Democratic leaders have been unable to reach agreement on a spending plan that is both ambitious and bipartisan.}\\
\hline
\textsc{Human}: \textit{A group of moderate House Democrats have initiated conversations with Republican lawmakers in a bid to try to reach a deal to tackle burgeoning federal spending. The group is being led by Reps. Jim Cooper (D., Tenn.) and Kurt Schrader (D., Ore.), both of whom are \textbf{leaders} of the Blue Dogs, a \textbf{coalition} of fiscal \textbf{conservative} Democrats. Tuesday, the \textbf{lawmakers} released a series of fiscal discipline \textbf{goals}, many of which are similar to the targets set in place by the deficit commission established...} 
\\
\hline
\hline
\end{tabularx}
\end{center}
\clearpage

\begin{center}
\begin{tabularx}{\linewidth}{X}
\hline
\hline
Text 4 \\ 
Context: \textcolor{blue}{Oscar nominated actress Hailee Steinfeld is set to play Sleeping Beauty in a re-visioning of the classic fairytale. The 14-year-old star has signed on to play the unfortunate princess in} \\ 
Keywords: movie, version, story, star, sleep, dream, role, novel, upcoming, youngest, actress, nominated, television, adaptation, classic \\
\hline
\textsc{Ours + Gpt2-774M}: \textit{Oscar nominated actress Hailee Steinfeld is set to play Sleeping Beauty in a re-visioning of the classic fairytale. The 14-year-old star has signed on to play the unfortunate princess in the \textbf{adaptation} of \textbf{classic} \textbf{novel} \textbf{Sleeping} Beauty \textbf{starring} \textbf{actress} Emma Watson. The \textbf{movie} will be directed by Story \textbf{television} and produced by \textbf{DreamWorks} Animation.
The \textbf{role} of Sleeping Beauty was originally played by Australian actress Sophie Okonedo in the original \textbf{version} of the \textbf{story}. The \textbf{upcoming} remake will be directed by Oscar \textbf{nominated} director Rob Marshall.
Steinfeld, who is the \textbf{youngest} of four siblings and a native of Melbourne, Australia, has already landed a pair of teen movies with her younger sister Zoe in Zoe and Me and Ice Age.
"It's exciting to be working with DreamWorks again, and on a film like Sleeping Beauty," Steinfeld said.}\\
\hline
\textsc{Human}: \textit{Oscar nominated actress Hailee Steinfeld is set to play Sleeping Beauty in a re-visioning of the classic fairytale. The 14-year-old star has signed on to play the unfortunate princess in the new \textbf{movie}. It will be a spin of a \textbf{version} of the \textbf{story} which is being written by screen writer Lindsay Devlin. According to Deadline.com the new version will give the \textbf{star} more to do than just \textbf{sleep} as it will follow her as she enters a \textbf{dream} world and has to find her way out. Hailee is also currently being considered to play the lead \textbf{role} in the \textbf{novel}-based film Forgotten, from the \textbf{upcoming} book of the same name by Cat Patrick. After being one of the \textbf{youngest} ever \textbf{actresses} to be \textbf{nominated} for an Academy Award at this years Oscars she is set to have a busy year with several other offers on the table. The young California native had only starred in several \textbf{television} bit parts before landing the lead role in the Cohen Brothers \textbf{adaptation} of the \textbf{classic} John Wayne western True Grit alongside Jeff Bridges.}
\\
\hline
\hline
\end{tabularx}
\end{center}

\begin{center}
\begin{tabularx}{\linewidth}{X}
\hline
\hline
Text 5 \\ 
Context: \textcolor{blue}{Pinch-hitter David Murphy delivered a tiebreaking, two-run double in the eighth inning and the Texas Rangers rallied for a 9-5 victory over the Boston Red Sox on Friday after raising} \\ 
Keywords: championship, kicked, land, newcomers, home, score, played, season, victory, hits, runs, debut, strikeouts \\
\hline
\textsc{Ours + Gpt2-774M}: \textit{Pinch-hitter David Murphy delivered a tiebreaking, two-run double in the eighth inning and the Texas Rangers rallied for a 9-5 victory over the Boston Red Sox on Friday after raising their \textbf{season} \textbf{home} run total to 27.
The \textbf{victory} was the Rangers' first since \textbf{hitting} 27 home \textbf{runs} in a season-opening win over the New York Yankees on April 1.
The Rangers scored four runs in the eighth inning to take a 6-5 lead. Murphy \textbf{played} the role of a pinch hitter in the eighth inning and \textbf{kicked} a single to right field to \textbf{score} Mark Trumbo and make it 7-5.
The Rangers scored four more runs in the ninth inning to \textbf{land} the win. Matt Bush hit a two-run homer to right field and Prince Fielder hit a two-run double to left field to make it 9-5. The \textbf{strikeouts} were the most the Rangers have allowed in a game since \textbf{debuting} with 26 in a win over the New York Yankees on April 1.
The \textbf{championship} of the American League West will be decided on Sunday when the Rangers host the \textbf{newcomers} the Los Angeles Angels.}\\
\hline
\textsc{Human}: \textit{Pinch-hitter David Murphy delivered a tiebreaking, two-run double in the eighth inning and the Texas Rangers rallied for a 9-5 victory over the Boston Red Sox on Friday after raising their American League \textbf{championship} flag. Murphy's slicing liner off Daniel Bard (0-1) \textbf{kicked} up chalk when it \textbf{landed} on the left-field line. That sent Rangers \textbf{newcomers} Mike Napoli and Yorvit Torrealba \textbf{home} to break a 5-all tie. Murphy \textbf{scored} on a double by Elvis Andrus before another double by AL MVP Josh Hamilton. Napoli, Ian Kinsler and Nelson Cruz all homered off Jon Lester for Texas, which \textbf{played} its \textbf{season} opener exactly five months after a Game 5 loss to San Francisco at home ended its first World Series. Darren Oliver (1-0) got the \textbf{victory} after allowing a tying homer to David Ortiz in the eighth. Adrian Gonzalez had two \textbf{hits} and drove in three \textbf{runs} in his Boston \textbf{debut.} Carl Crawford went 0 for 4 with three \textbf{strikeouts} while leaving a runner in scoring position each at-bat.} 
\\
\hline
\hline
\end{tabularx}
\end{center}

\begin{center}
\begin{tabularx}{\linewidth}{X}
\hline
\hline
Text 6 \\ 
Context: \textcolor{blue}{An Alaska lawmaker introduced an amendment Friday that would give oil companies a tax break provided they pledged to increase production. Alaska Gov. Sean Parnell, a Republican, unveiled plans for} \\ 
Keywords: barrels, production, pipeline, companies, investments, oil, tax, message, legislators, strategy, future, amendment, restrictions, companies, tax, increase, percentage, benefit \\
\hline
\textsc{Ours + Gpt2-774M}: \textit{An Alaska lawmaker introduced an amendment Friday that would give oil companies a tax break provided they pledged to increase production. Alaska Gov. Sean Parnell, a Republican, unveiled plans for a \textbf{tax} credit for \textbf{oil} companies that \textbf{increase} \textbf{production}. \textbf{Pipeline} \textbf{companies} would \textbf{benefit} from \textbf{tax} breaks as well.
The \textbf{amendment} would \textbf{restrict} oil \textbf{companies} from receiving a tax credit if they increase production by more than 10 percent a year for five years and \textbf{invest} more than \$1 billion in \textbf{future} projects. The \textbf{legislation} is expected to be introduced in the House and Senate this week.
Parnell said the \textbf{percentage} increase would be based on the amount of \textbf{barrels} of oil produced by the company.
"This is a great opportunity for us to help the oil industry grow and create jobs," Parnell said.
The \textbf{strategy} is expected to be controversial in Alaska, where the oil industry has been a major employer and a major contributor to the state's economy. The state has the nation's third-largest oil reserves and is home to the world's largest crude oil \textbf{messageboard}.}\\
\hline
\textsc{Human}: \textit{An Alaska lawmaker introduced an amendment Friday that would give oil companies a tax break provided they pledged to increase production. Alaska Gov. Sean Parnell, a Republican, unveiled plans for 1 million \textbf{barrels} of oil \textbf{production} per day through the Trans-Alaska \textbf{pipeline} system within the next decade. Parnell said he was proposing a tax break for oil \textbf{companies} to encourage \textbf{investments} and to erase declines in state \textbf{oil} production. "The time to reduce oil \textbf{taxes} is now and I am asking all Alaskans to send a clear \textbf{message} to \textbf{legislators} in Juneau that a "do-nothing" \textbf{strategy} is unacceptable because Alaska\'s \textbf{future} is at stake," he said in a statement. But state Rep. Bob Miller, D-Fairbanks, introduced an \textbf{amendment} that would put certain \textbf{restrictions} on oil \textbf{companies.} Under Miller\'s plan, companies would get a \textbf{tax} break until January 2017. After that, they would need to have \textbf{increased} production by 10 percent of current levels to continue getting a break and increase production by another 2 \textbf{percent} each year beyond 2017. "We want to be sure that they are earning the breaks that we are giving," Miller said in a statement. "This amendment says here\'s the \textbf{benefit}. If you do not make certain metrics for the benefit of Alaska, we withdraw those benefits."} 
\\
\hline
\hline
\end{tabularx}
\end{center}

\begin{center}
\begin{tabularx}{\linewidth}{X}
\hline
\hline
Text 7 \\ 
Context: \textcolor{blue}{At a government laboratory in Alabama, workers in blue coats unload envelopes packed with small filters that trapped air particles in Hawaii, Alaska and elsewhere. The discs are placed in} \\ 
Keywords: devices, traces, radioactive, materials, ocean, dangerous, workers, disaster, radiation, environmental, administration, federal, consumers, crisis, water \\
\hline
\textsc{Ours + Gpt2-774M}: \textit{At a government laboratory in Alabama, workers in blue coats unload envelopes packed with small filters that trapped air particles in Hawaii, Alaska and elsewhere. The discs are placed in a \textbf{water}-filled \textbf{device} called a " federal \textbf{environmental} control and \textbf{radiation} protection system," or E \textbf{consumers}, which \textbf{traces} \textbf{radioactive} \textbf{material} to the \textbf{ocean} floor. The \textbf{workers} put the discs in a machine that separates the \textbf{dangerous} particles from the rest of the water.
The process is called " \textbf{crisis} management " and is used to deal with disasters such as the Fukushima nuclear \textbf{disaster} in Japan in 2011. The E consumers are also used to deal with the fallout from the Chernobyl nuclear disaster in 1986 in the Ukraine.
The \textbf{federal} government has spent \$5 billion on the system since the \textbf{administration} of President George W. Bush signed the U.S. Environmental Protection Agency Reactor Resilience Initiative, a 35-year program that aims to turn the Fukushima Daiichi nuclear plant's reactors into low-carbon and efficient power plants. It has provided \$1 billion each year since it was established in the aftermath of the tragedy.
If each E consumer is turned on once a year for 40 years, a certain percentage of radioactivity on the Pacific Ocean floor will be removed. This includes particulates in the ocean bottom, as well as radioactive materials such as uranium. It also includes contaminated groundwater.}\\
\hline
\textsc{Human}: \textit{At a government laboratory in Alabama, workers in blue coats unload envelopes packed with small filters that trapped air particles in Hawaii, Alaska and elsewhere. The discs are placed in lead-lined, barrellike \textbf{devices} for testing to make sure no \textbf{traces} of \textbf{radioactive} \textbf{materials} have wafted across the Pacific \textbf{Ocean} from Japan. So far, the sea breeze in places such as Honolulu is no more \textbf{dangerous} than the pollen-laden air of the Deep South, according to officials. Still, the 60 or so \textbf{workers} in the 72,000-square-foot building will be the first to know if the Japanese \textbf{disaster} spreads harmful amounts of \textbf{radiation} to the U.S. On Wednesday, the \textbf{Environmental} Protection Agency and the Food and Drug \textbf{Administration} said very low levels of radiation had turned up in a sample of milk from Washington state, but \textbf{federal} officials assured \textbf{consumers} not to worry. The FDA said such findings were to be expected in the coming days because of the nuclear \textbf{crisis} in Japan and that the levels were expected to drop relatively quickly. The EPA said it was increasing the level of nationwide monitoring of milk, precipitation and drinking \textbf{water}.} 
\\
\hline
\hline
\end{tabularx}
\end{center}

\begin{center}
\begin{tabularx}{\linewidth}{X}
\hline
\hline
Text 8 \\ 
Context: \textcolor{blue}{A 25-year-old German man has been arrested for allegedly burying a cache of bombs near a German soccer stadium in a blackmail plot, authorities say. The unnamed German national was} \\ 
Keywords: german, arrested, explosives, garage, stadium, federal, criminal, investigation, bombs, home, embassy, information, blackmail, criminal, terrorist, organizations, safe \\
\hline
\textsc{Ours + Gpt2-774M}: \textit{A 25-year-old German man has been arrested for allegedly burying a cache of bombs near a German soccer stadium in a blackmail plot, authorities say. The unnamed German national was \textbf{arrested} in the \textbf{German} city of Leipzig on Monday, \textbf{federal} prosecutors said. \textbf{Investigators} believe the \textbf{explosives} were hidden in a \textbf{home} \textbf{garage} and \textbf{bombs} were hidden in a safe in the garage, according to the \textbf{Criminal} \textbf{Criminal} Information Service.
The suspect is suspected of blackmailing a German soccer \textbf{stadium} security guard and threatening to detonate a \textbf{terrorist} attack if he didn't comply with his demands, the prosecutors said. The suspect is also suspected of trying to obtain \textbf{information} about organizations and individuals involved in the soccer stadium security.
The suspect is suspected of \textbf{blackmailing} a German soccer stadium security guard and threatening to detonate a terrorist attack if he didn't comply with his demands, the prosecutors said. The suspect is also suspected of trying to obtain information about organizations and individuals involved in the soccer stadium security. \textbf{Embassy} Security Blog discusses the "Security Update".}\\
\hline
\textsc{Human}: \textit{A 25-year-old German man has been arrested for allegedly burying a cache of bombs near a German soccer stadium in a blackmail plot, authorities say. The unnamed \textbf{German} national was \textbf{arrested} in Cologne on Tuesday after allegedly placing the \textbf{explosives} in a parking \textbf{garage} near the \textbf{Westfalenstadion} in Dortmund, home of the Borussia Dortmund team, the \textbf{Federal} Office of \textbf{Criminal} \textbf{Investigation} told The Local news agency. The \textbf{bombs} were safely defused, and three more were found at the man's \textbf{home} in Krefeld, officials said. Investigators said they began tracking the man after he e-mailed the German \textbf{Embassy} in Pakistan, offering \textbf{information} about two planned attacks in Germany by a group. The warning appeared to be a \textbf{blackmail} bid and was worded like an unsolved attempted blackmail case last year. "The suspect apparently acted alone with a general \textbf{criminal} motive," a federal spokesman said. "There are absolutely no ties to \textbf{terrorist} or Islamist \textbf{organizations."} Authorities say he admitted placing the bombs. Dortmund police spokesman Michael Stein told the BBC: "We expect no security threat at all for the upcoming Bundesliga match on Saturday. Visitors are invited to come to Dortmund. They will be \textbf{safe} here."} 
\\
\hline
\hline
\end{tabularx}
\end{center}

\begin{center}
\begin{tabularx}{\linewidth}{X}
\hline
\hline
Text 9 \\ 
Context: \textcolor{blue}{I’d enjoy reading the citation of constitutional authority for this: “The bill then says if the Senate does not act, then H.R. 1 [the House-passed bill that cuts \$61 billion]} \\ 
Keywords: senate, shutdown, pay, unconstitutional, prevention, legislation, president, sign, law, senate, compromise, problem, effort, attention, deal, option \\
\hline
\textsc{Ours + Gpt2-774M}: \textit{I’d enjoy reading the citation of constitutional authority for this: “The bill then says if the Senate does not act, then H.R. 1 [the House-passed bill that cuts \$61 billion] will become \textbf{law} without \textbf{Senate} action. This \textbf{legislation} is \textbf{unconstitutional} because \textbf{Senate} \textbf{President} Pro Tempore (and the \textbf{signers} of the Constitution) have the \textbf{option} of not passing it. This \textbf{problem} does not exist in the House bill.
This is the same \textbf{effort} that has been going on for years to get the Senate to act on the \textbf{shutdown} \textbf{deal}, and it has not worked. The \textbf{compromise} that was reached last week was not enough to get the \textbf{attention} of the Senate to act on the bill. So now they are \textbf{paying} the price for not acting \textbf{prevention} of a government shutdown happening, which all sides recognize will happen, eventually.
}\\
\hline
\textsc{Human}: \textit{I’d enjoy reading the citation of constitutional authority for this: “The bill then says if the Senate does not act, then H.R. 1 [the House-passed bill that cuts \$61 billion] will be the law of the land. In addition to that, it says that if all else fails, and the \textbf{Senate} brings about a \textbf{shutdown}, then members should not get their \textbf{pay}.” That’s House Majority Leader Eric Cantor describing his “Government Shutdown \textbf{Prevention} Act.” The problem is, this would be blatantly \textbf{unconstitutional}: The Senate needs to pass the same piece of \textbf{legislation} the House does, and the president needs to either \textbf{sign} it or have his veto overturned. That’s how deem-and-pass worked with the health-care \textbf{law}, for instance: Both the Senate and the House passed the same pieces of legislation, and then the \textbf{president} signed them. But it seems Cantor merely misspoke. I’ve clarified with both his office and Boehner’s office that they believe the \textbf{Senate} and the president would still play their traditional roles. That means deem-and-pass isn't, as Cantor suggests, an alternative to actually striking a \textbf{compromise}. It’s just an \textbf{effort} to message the shutdown that’ll happen if a law isn’t passed. The \textbf{problem} for Cantor is that by misdescribing how the gambit would work, he’s drawing \textbf{attention} to the fact that it can’t. At the end of the day, we need an actual \textbf{deal} here. There’s no other \textbf{option}.} 
\\
\hline
\hline
\end{tabularx}
\end{center}

\begin{center}
\begin{tabularx}{\linewidth}{X}
\hline
\hline
Text 10 \\ 
Context: \textcolor{blue}{We tend to think of Seattle in stereotypical ways: earthy, mellow, panoramic, rainy. But it's not all Patagonia and lattes out there, as the news yesterday about the Justice Department's} \\ 
Keywords: investigation, police, force, discriminating, yesterday, robbery, suspect, detective, african, native, knife, chief, investigation, justice \\
\hline
\textsc{Ours + Gpt2-774M}: \textit{We tend to think of Seattle in stereotypical ways: earthy, mellow, panoramic, rainy. But it's not all Patagonia and lattes out there, as the news yesterday about the Justice Department's \textbf{investigation} of \textbf{police} \textbf{force} in Seattle shows. The \textbf{Justice} Department investigation, which was announced \textbf{yesterday}, is focused on the \textbf{Chief} of Police, who is African American, and the Seattle Police Department, which is \textbf{Native} American.
The Seattle Police Department has been under federal \textbf{investigation} for a long time. The Justice Department has \textbf{suspected} that the Seattle Police Department has \textbf{discriminated} against \textbf{African} Americans and other minorities. The \textbf{detective} in charge of the investigation, who is white, is a member of the department's Professional Standards Bureau.
The investigation has been going on for a long time. The \textbf{robbery} of a black man in Seattle in the early 1990s was one of the first cases that the Professional Standards Bureau investigated. The detective in charge of the investigation, who is white, was a \textbf{knife}-wielding undercover officer. There have been other violent incidents involving white Seattle police officers and Asian men.}\\
\hline
\textsc{Human}: \textit{We tend to think of Seattle in stereotypical ways: earthy, mellow, panoramic, rainy. But it's not all Patagonia and lattes out there, as the news yesterday about the Justice Department's \textbf{investigation} of Seattle \textbf{police} would indicate. The DOJ is looking into a possible pattern of the SPD using excessive \textbf{force} and \textbf{discriminating} against minorities, Justice announced here \textbf{yesterday}. At issue in the federal investigation are several high-profile incidents involving police violence. Last April, a \textbf{detective} was videotaped kicking a Latino \textbf{robbery} \textbf{suspect} and stating that he would beat the "Mexican piss" out of the suspect; In June, an officer was videotaped punching a 17-year-old \textbf{African} American girl who protested the arrest of a friend for jaywalking; and, In August, police shot a \textbf{Native} American woodcarver after he faied to drop his carving \textbf{knife}. Seattle Police \textbf{Chief} John Diaz told the Seattle Times that he welcomes the DOJ \textbf{investigation} and considers it like a "free audit." "We have nothing to hide," he said. "We've been open and transparent with the Department of \textbf{Justice}, which makes for a good working relationship."} 
\\
\hline
\hline
\end{tabularx}
\end{center}

\clearpage
\subsection{Detailed Results by News Article}
\label{app:news_results}

\begin{figure}
    \centering
    \includegraphics[width=\linewidth]{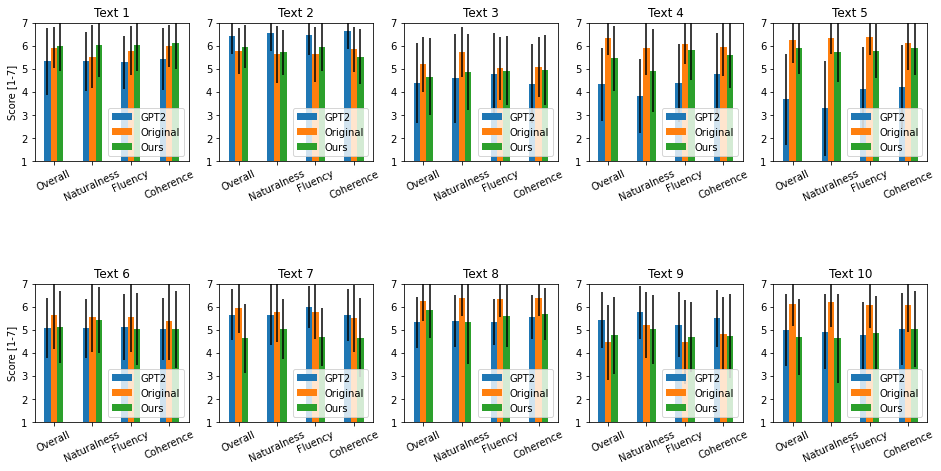}
    \caption{Results of the user study with mean scores and standard deviation error bars per text.}
    \label{fig:news_results_by_q}
\end{figure}

In Figure \ref{fig:news_results_by_q} we show the results of the user study split by news article. As with the aggregated results, we see that per article the average scores for our method are very similar to the scores for uncontrolled GPT-2. However, there are some noteworthy differences. In particular, in the cases of texts $4$ and $5$, our method scores much higher than uncontrolled GPT-2 on all four scales. On the other hand uncontrolled GPT-2 performs exceptionally well on text $2$, but without meeting the keyword requirements of course. 

Although on average our method lags behind human-written text in all four categories, it performs almost on par for many of the articles, and even produces text of higher perceived quality than the original for the first article. On the other hand our method struggles to match the scores in all categories for text $10$ and texts $7$ and $8$ to a lesser extent. 

There is not much variation across the four categories even in the detailed results.

\end{document}